%% file: main.tex
\documentclass[sigconf]{acmart}
\input{math_commands.tex}

\usepackage{hyperref}
\usepackage{url}
\usepackage{booktabs}       
\usepackage{amsfonts}       
\usepackage{nicefrac}       
\usepackage{microtype}      
\usepackage{subcaption}
\usepackage{multirow}

\usepackage{graphicx}
\usepackage{comment}
\usepackage{amsmath}
\usepackage{color}
\usepackage{paralist}
\usepackage{bm}
\usepackage{pifont}

\newcommand{\lu}[1]{{\textcolor{red}{[lu: #1]}}}

\input{macros.tex}

\newcommand{\fancynamenospace}{TIM-GAN}
\newcommand{\fancyname}{\fancynamenospace~}

\newcommand{\MLP}{f_{\mathrm{MLP}}}

\newcommand{\phiwhere}{\phi^\mathrm{where}_t}
\newcommand{\phihow}{\phi^\mathrm{how}_t}
\newcommand{\phiout}{\phi_{\hat{y}}}

\newcommand{\fhow}{f_{\mathrm{how}}}
\newcommand{\fwhere}{f_{\mathrm{where}}}
\newcommand{\fop}{f_{\mathrm{how}}}

\newcommand{\thetat}{\Theta_{\mathrm{how}}(t)}

\AtBeginDocument{%
  \providecommand\BibTeX{{%
    \normalfont B\kern-0.5em{\scshape i\kern-0.25em b}\kern-0.8em\TeX}}}

\setcopyright{rightsretained}

\acmSubmissionID{1013}

\begin{document}
\fancyhead{}
\title{Text as Neural Operator: Image Manipulation by Text Instruction}

\copyrightyear{2021}
\acmYear{2021}
\acmConference[MM '21]{Proceedings of the 29th ACM International Conference on Multimedia}{October 20--24, 2021}{Virtual Event, China}
\acmBooktitle{Proceedings of the 29th ACM International Conference on Multimedia (MM '21), October 20--24, 2021, Virtual Event, China}\acmDOI{10.1145/3474085.3475343}
\acmISBN{978-1-4503-8651-7/21/10}
\author{Tianhao Zhang}
\authornote{Work done as a Google AI Resident.}
\affiliation{%
  \institution{Google Research}
  \country{}
  }
\email{bryanzhang@google.com}

\author{Hung-Yu Tseng}
\authornote{Work done during HY’s internship at Google Research.}
\affiliation{%
  \institution{University of California, Merced}
  \country{}
  }
\email{htseng6@ucmerced.edu}

\author{Lu Jiang}
\affiliation{%
  \institution{Google Research}
  \country{}
  }
\affiliation{%
  \institution{Carnegie Mellon University}
  \country{}
  }
\email{lujiang@google.com}

\author{Weilong Yang}
\affiliation{%
  \institution{Waymo}
  \country{}
  }
\email{weilongyang@google.com}

\author{Honglak Lee}
\affiliation{%
  \institution{University of Michigan}
  \country{}
  }
\email{honglak@eecs.umich.edu}

\author{Irfan Essa}
\affiliation{%
  \institution{Google Research}
  \country{}
  }
\affiliation{%
  \institution{Georgia Institute of Technology}
  \country{}
  }
\email{irfanessa@google.com}


\input{0_abstract.tex}



\begin{teaserfigure}
    \centering
    \vspace{-3mm}
    \includegraphics[width=\linewidth]{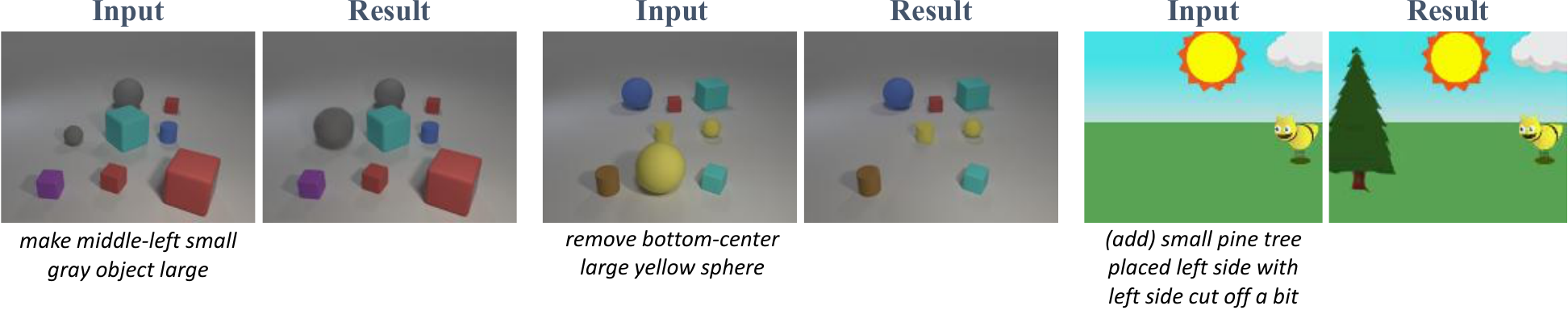}
    \vspace{-8mm}
     \caption{\textbf{Image manipulation by text instruction.} The input is multimodal consisting of a reference image and a text instruction. The results are synthesized images by our model.}
    \label{fig:teaser}
\end{teaserfigure}

\maketitle
\input{1_introduction.tex}
\input{2_related.tex}
\input{3_method.tex}

\input{4_experiments.tex}

\input{5_conclusion.tex}

\clearpage
\balance
\bibliographystyle{ACM-Reference-Format}
\bibliography{citation}



\end{document}


\title{Text as Neural Operator: Image Manipulation by Text Instruction Supplementary Material}

\copyrightyear{2021}
\acmYear{2021}
\acmConference[MM '21]{Proceedings of the 29th ACM International Conference on Multimedia}{October 20--24, 2021}{Virtual Event, China}
\acmBooktitle{Proceedings of the 29th ACM International Conference on Multimedia (MM '21), October 20--24, 2021, Virtual Event, China}\acmDOI{10.1145/3474085.3475343}
\acmISBN{978-1-4503-8651-7/21/10}
\author{Tianhao Zhang}
\authornote{Work done as a Google AI Resident.}
\affiliation{%
  \institution{Google Research}
  \country{}
  }
\email{bryanzhang@google.com}

\author{Hung-Yu Tseng}
\authornote{Work done during HY’s internship at Google Research.}
\affiliation{%
  \institution{University of California, Merced}
  \country{}
  }
\email{htseng6@ucmerced.edu}

\author{Lu Jiang}
\affiliation{%
  \institution{Google Research}
  \country{}
  }
\affiliation{%
  \institution{Carnegie Mellon University}
  \country{}
  }
\email{lujiang@google.com}

\author{Weilong Yang}
\affiliation{%
  \institution{Waymo}
  \country{}
  }
\email{weilongyang@google.com}

\author{Honglak Lee}
\affiliation{%
  \institution{University of Michigan}
  \country{}
  }
\email{honglak@eecs.umich.edu}

\author{Irfan Essa}
\affiliation{%
  \institution{Google Research}
  \country{}
  }
\affiliation{%
  \institution{Georgia Institute of Technology}
  \country{}
  }
\email{irfanessa@google.com}


\maketitle
\section{More Qualitative Results}

\subsection{Images generated by our model}
\begin{figure*}[ht]
    \centering
    \includegraphics[width=0.9\linewidth]{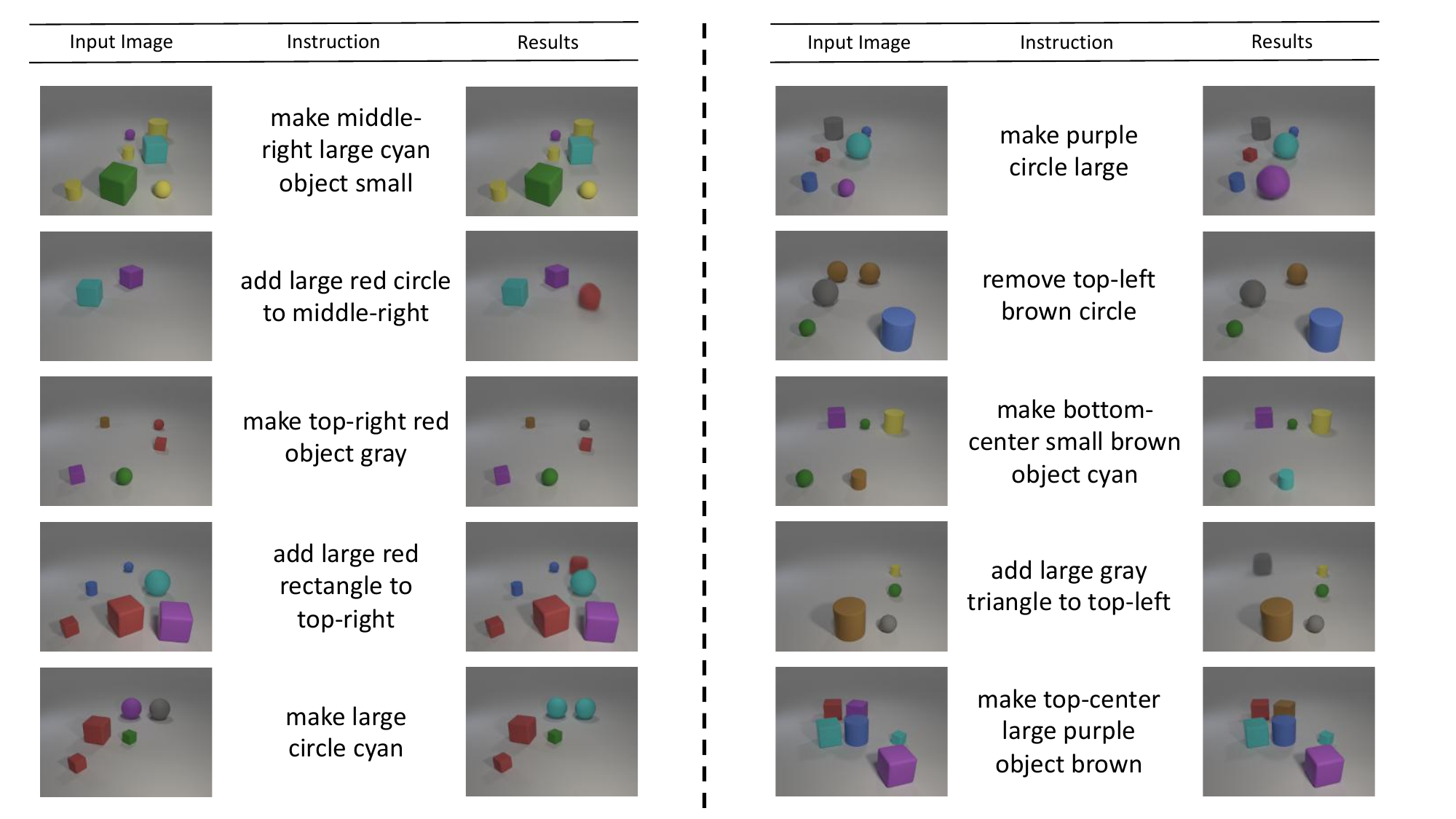}
    \caption{\textbf{Examples of the generated image by our model on Clevr.} }
    \label{fig:moreclevr}
\end{figure*}

\begin{figure*}[ht]
    \centering
    \includegraphics[width=0.9\linewidth]{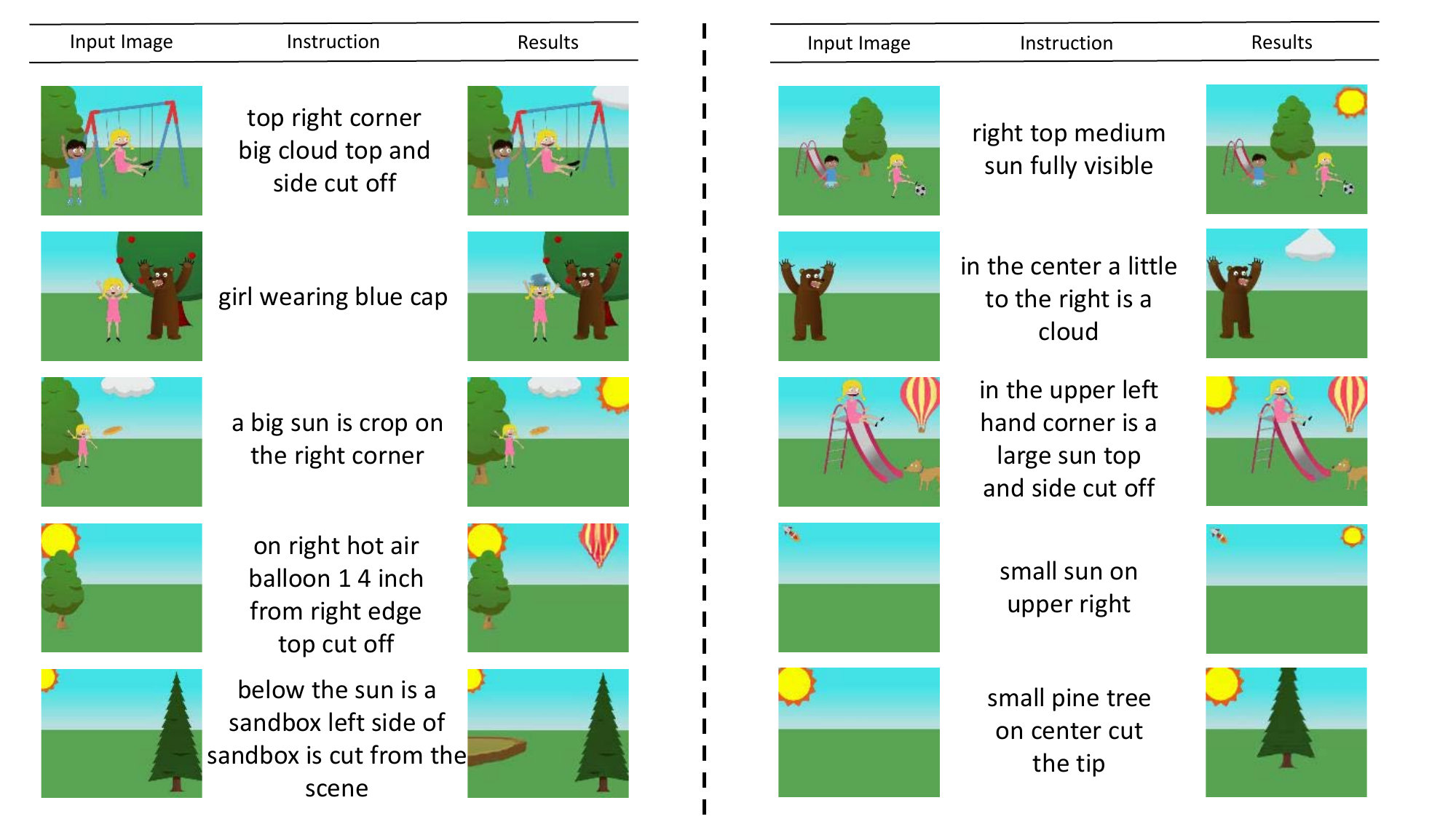}
    \caption{\textbf{Examples of the generated image by our model on Abstract Scene.} }
    \label{fig:moreabstract}
\end{figure*}

\begin{figure*}[ht]
    \centering
    \includegraphics[width=0.9\linewidth]{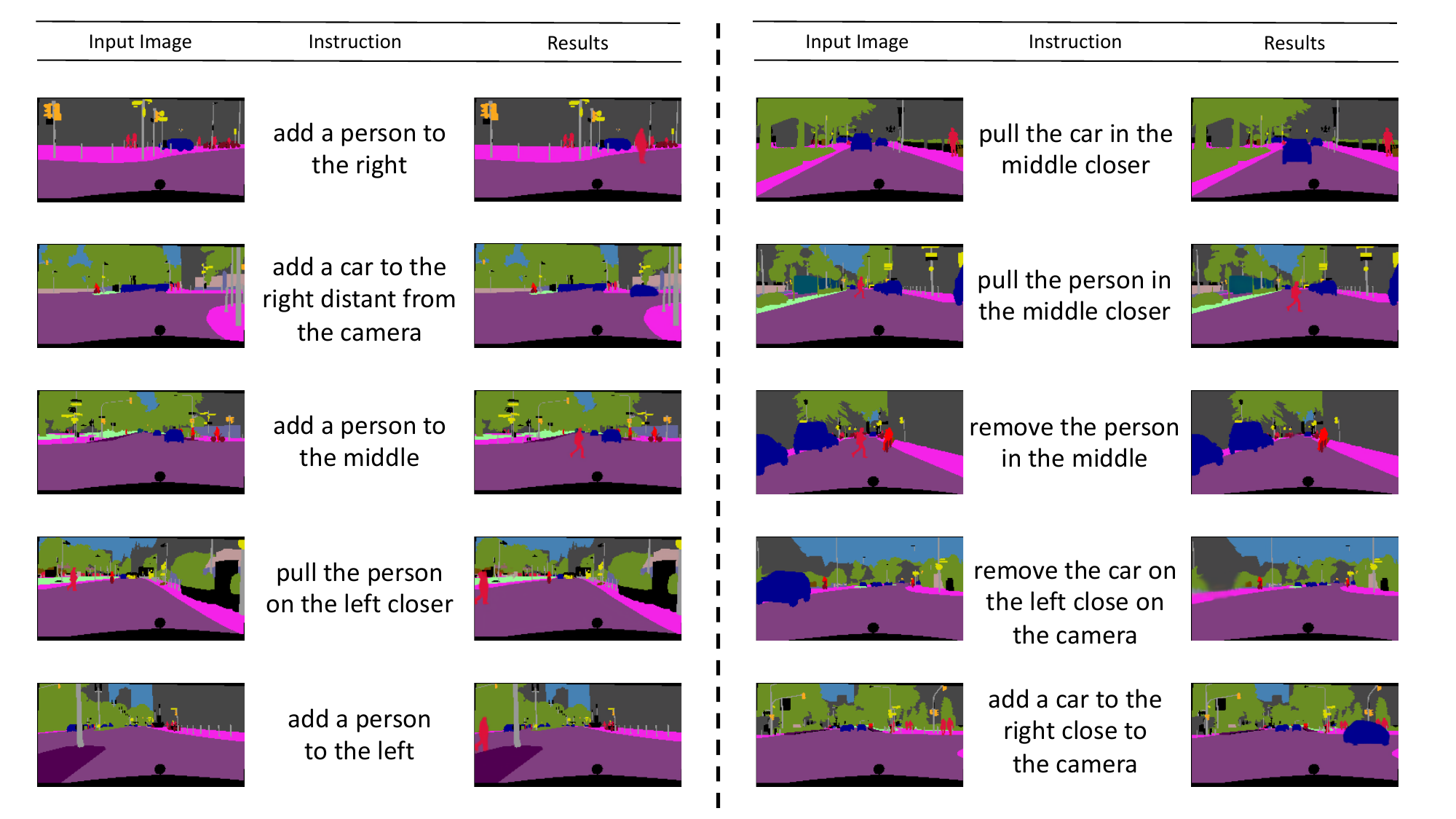}
    \caption{\textbf{Examples of the generated image by our model on Cityscapes.} }
    \label{fig:morecity}
\end{figure*}

\begin{figure*}[ht]
    \centering
    \includegraphics[width=\linewidth]{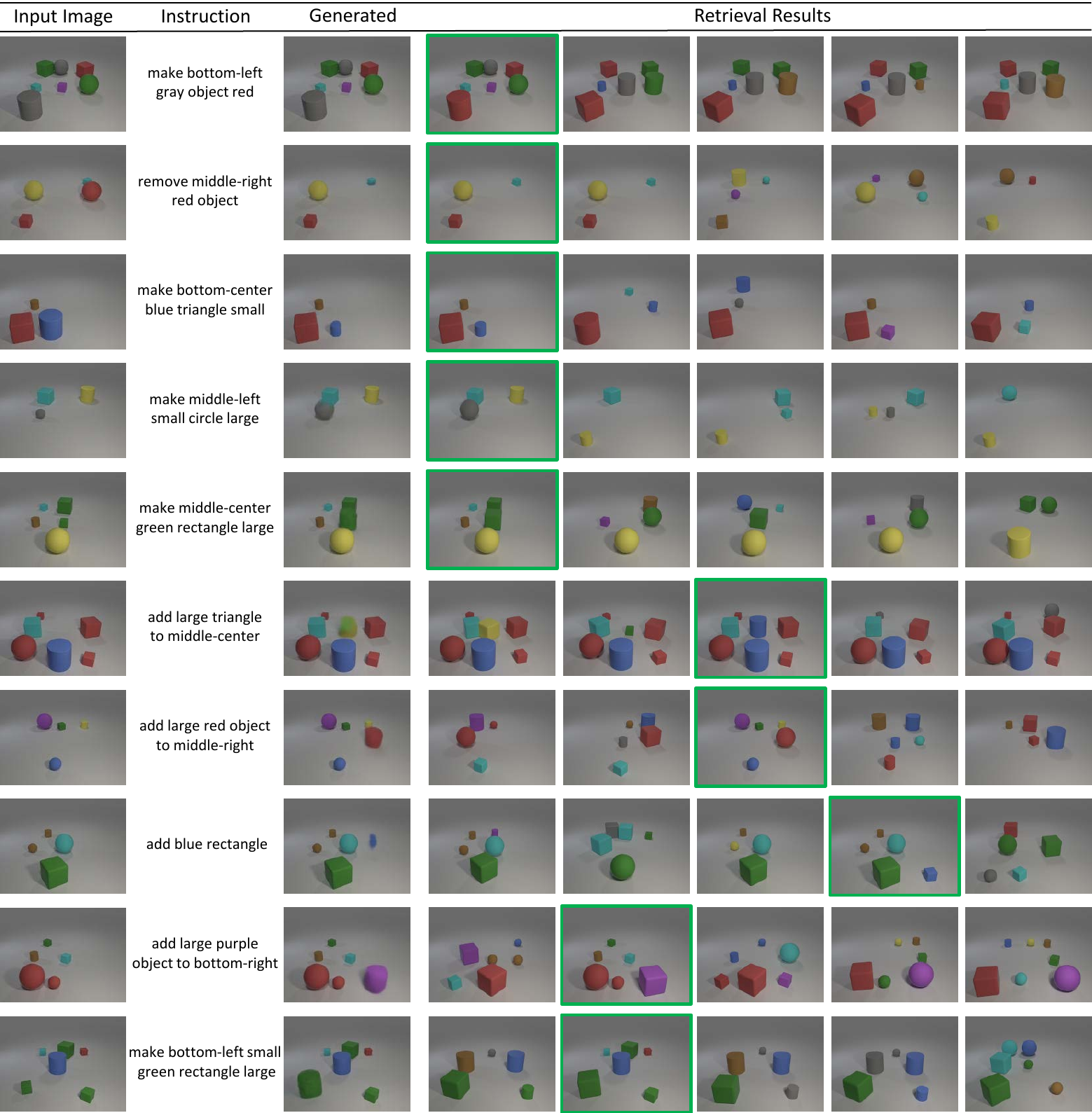}
    \caption{\textbf{Retrieval Results.} For each row, top-5 retrieved images are shown. The correct image is highlighted in the green box.}
    \label{fig:retrieval}
\end{figure*}

We show additional images generated by our model on the three experimental datasets. See Figure~\ref{fig:moreclevr}, Figure~\ref{fig:moreabstract}, and Figure~\ref{fig:morecity} for details. Generally, our model can handle complex text instructions. But we also observe cases in which our method can fail: (a) when the location of the target is not well-specified, see Figure~\ref{fig:retrieval} the $8$-th row; (b) when the attribute of the target is not detailed enough, see Figure~\ref{fig:retrieval} the $7$-th and $9$-th row.

\subsection{Retrieval results}
We use the generated image by our model as a query to retrieve the target image. Figure~\ref{fig:retrieval} shows the top-5 retrieved images on the Clevr dataset.
We show the successful retrieval cases in the first 5 rows and failure cases in the rest of the  rows. 




\section{Data Processing Details}
\subsection{Clevr}
We use the CSS dataset~\citep{vo2019tirg} which was created for the image retrieval task. 
The dataset is generated using the Clevr toolkit~\citep{johnson2017clevr} and contains 3-D synthesized images with the presence of objects with different color, shape, and size. 
Each training sample includes an input image, an output image and a text instruction specifying the modification. 
There are three types of modifications: add a new object, remove an existing object, and change the attribute of an object. 
Each text instruction specifies the position of the target object and the expected modification.
The dataset includes 17K training data pairs and 17K tests.
Note the original rendering of this dataset contains significant camera and object displacements which fail GAN model training of all the methods. In our experiments, we use the official raw dataset obtained from the authors of~\cite{vo2019tirg} and re-render the images to reduce the misalignment for unchanged objects. As a result, we can train meaningful GAN models and compare all methods fairly on the same CSS benchmark.

\subsection{Abstract scene}
CoDraw\citep{kim2017codraw} is a synthetic dataset built upon the Abstract Scene dataset\citep{zitnick2013bringing}. 
It is formed by sequences of images of children playing in the park. 
For each sequence, there is a conversation between a Teller and a Drawer. %
The teller gives text instructions on how to change the current image and the Drawer can ask questions to confirm details and output images step by step. 
To adapt it to our setting, we extract the image and text of a single step.
The dataset consists of 30K training and 8K test instances. Each training sample includes an input image, an output image and a text description about the object to be added to the input image.

\subsection{Cityscapes}
We create a third dataset based on Cityscapes segmentation masks.
The dataset consists of 4 types of text modifications: ``add'', ``remove'', ``pull an object closer'', and ``push an object away''. The ground-truth images are manually generated by pasting desired objects on the input image at appropriate positions. We crop out various object prototypes (cars, people, etc.) from existing images. Specifically, adding is done by simply pasting the added object. Removing is the inverse of adding. Pulling and pushing objects are done by pasting the same object of different sizes (with some adjustment on location as well to simulate depth changing effect).
The dataset consists of 20K training instances and 3K examples for testing.

\section{Comparison with related Tasks}

\begin{figure*}[]
    \centering
    \includegraphics[width=\linewidth]{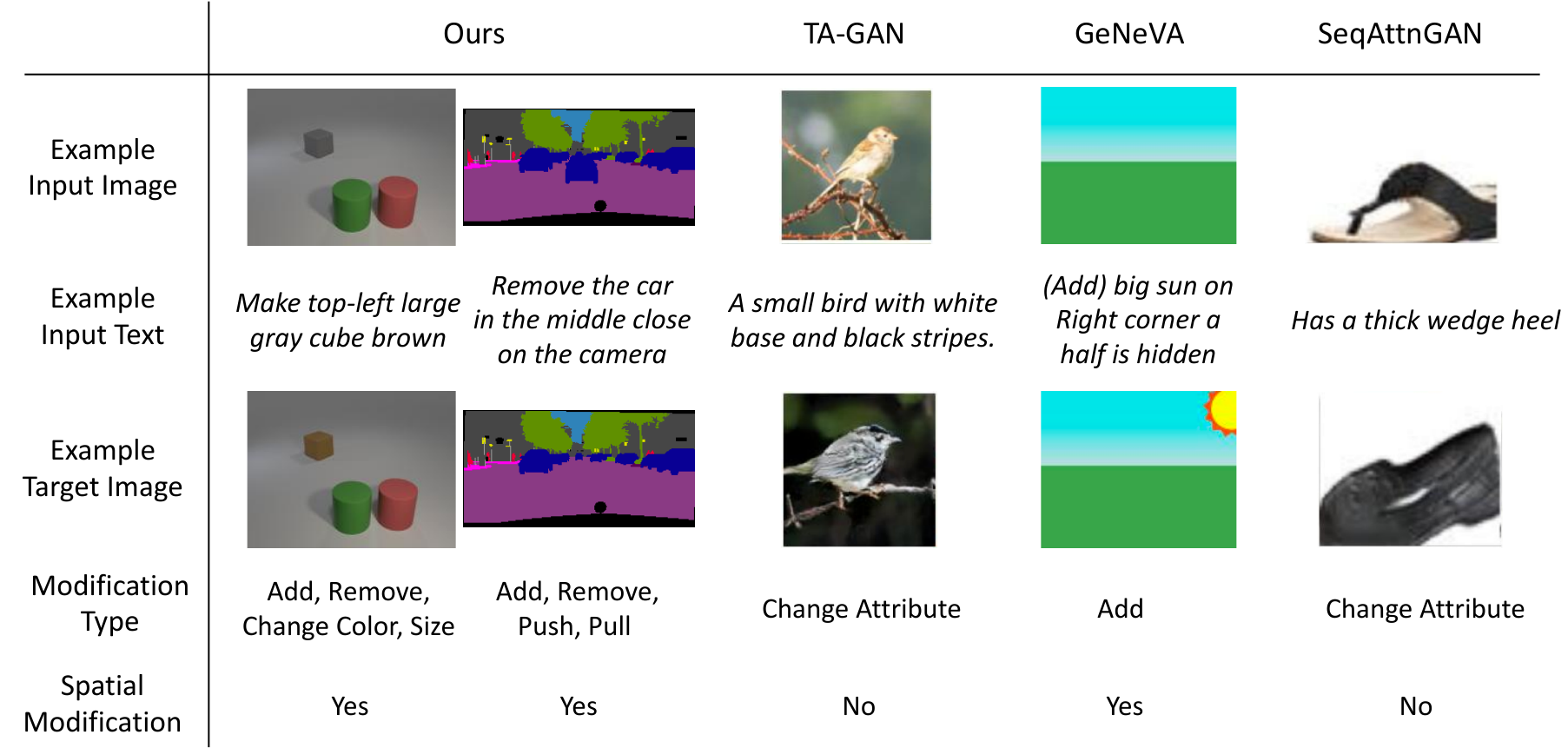}
    \caption{\label{fig:compare}\textbf{Comparison of different text-guided image manipulation settings.} }
\end{figure*}

This section discusses the differences between our work and the related works of image manipulation. It also explains the rationale of the synthetic dataset used in closely-related works.

\figref{compare} illustrates the comparison with three closely-related works: TA-GAN~\citep{nam2018tagan}, GeNeVA~\citep{elnouby2019geneva}, and SeqAttnGAN~\citep{cheng2020sequential}. Obviously, a commonality among them is that they share the same input format (a reference image and a text) and the output format (a target image).

Our work differs considerably from prior works in the task and evaluation dataset. In terms of the task, our work focuses on the \emph{complex text instructions} which cover three representative operations ``add'', ``modify'', and ``remove'', involving adjectives (attributes), verbs (actions) and adverbs (locations). On the other hand, the text in existing works are limited in complexity and diversity. For example, the captions in TA-GAN~\citep{nam2018tagan} or the descriptions in SeqAttnGAN~\citep{cheng2020sequential} mainly describe the attribute(s) to be changed without explicit spatial information. GeNeVA~\citep{elnouby2019geneva} is more relevant but it only performs a single type of operation~\ie adding objects.

The datasets are also different. The training data in SeqAttnGAN~\citep{cheng2020sequential} cannot be used for our task as the modification is less precise. As shown in \figref{compare}, the text instruction ``has a thick wedge heel'' turns a sandal into a leather shoe. In fact, it is difficult to a pair of real-world images with their associated change specified as texts. 
Due to the missing of suitable datasets, GeNeVA~\citep{elnouby2019geneva} and others~\citep{li2019storygan} generate the training and evaluation dataset in a simulation environment and were able to test on these synthetic images. However, the role of synthetic dataset in this research area is crucial as the synthetic dataset allows for not only an in-depth study (with controlled variables) but also a clearer evaluation of the generated results.
In this work, we expand it one step further to manipulate semantic segmentation in the real-world Cityscapes dataset. By doing so, we demonstrate the potential of our method for synthesizing RGB images from the modified segmentation mask.

It is worth noting the unsupervised learning approach~\cite{nam2018tagan} can learn image manipulation model without needing the true target image during training, These methods were mainly about captions but currently expanded it to text commands (attributes) by Liu~\etal~\citep{liu2020describe}.
Since these works address a separate challenge of unsupervised learning, we plan to study it as our future work. Nevertheless, we still need to collect a test set of parallel triples of the reference image, target image, and text instruction for a faithful evaluation.

\section{Experimental Details}

\subsection{Evaluation Details}
\paragraph{FID} We employ the standard FID~\citep{fid} metric based on the InceptionV3 model for the Clevr and the Abstract Scene datasets. On Cityscapes, the FID scores are computed using a pretrained auto-encoder on the semantic segmentation mask. 
We use the encoder to extract features for distance computation, and keep the feature dimension to be the same as the original Inception V3 network to produce the final score of a similar scale.

\paragraph{Retrieval Score} First, we extract the features of the edited images using the learned image encoder $E_i$ to get the queries.
For each query, we take the ground-truth output image and randomly select 999 real images from the test set, and then extract the features of these images using the same model to form a pool for the retrieval task.
Second, we compute the cosine similarity between the queries and image features from the pool.
We then select the top-$N$ most relevant images from the pool as the candidate set for each query.
%
%
We report Recall@$1$ and Recall@$5$ scores in our experiments, in which Recall@$N$ indicates the recall of the ground-truth image in the top-$N$ retrieved images.


\subsection{Implementation Details}
We implement our model in Pytorch~\citep{paszke2017pytorch}.
For the image encoder $E_i$, we use three down-sampling convolutional layers followed by Instance Normalization and ReLU activation. We use 3x3 kernels and a stride of 2 for down-sampling convolutional layers.
We construct the generator $G$ by using two residual blocks followed by three up-sampling layers (transposed-convolutional layers) followed by Instance Normalization and ReLU activation. We use 3x3 kernels and a stride of 2 for up-sampling layers.

As for the text encoder $E_t$, we use the BERT~\citep{devlin2018bert} model, specifically the cased version of \textit{BERT-Base}~\citep{devlin2018bert}. The parameters are initialized by their pretrained values.
%
%

The parameters in the image encoder $E_i$ and generator $G$ are initialized by training an image autoencoder. Specifically, for each dataset, we pre-train the image encoder and generator on all images of the dataset.
After the initialization, we fix the parameters in the image encoder $E_i$ and optimize the other parts of the network in the end-to-end training.
During pretraining of the autoencoder, we use the Adam optimizer~\citep{kingma2014adam} with a batch size of $8$, a learning rate of $0.002$, and exponential rates of $(\beta_1, \beta_2)=(0.5, 0.999)$ and train the model for 30 epochs.


%
The encoded image feature has $256$ channels. The BERT outputs text embeddings of dimension $d_0 = 768$, The dimension of the attended text embedding is $d = 512$. By default we use the routing-neurons strategy where the routing network has $l=2$ layers and $m=3$ blocks for each layer. For the sharing-neurons strategy used in our ablation studies,  we use the same number of layer $l=2$ but reduce $m$ to 1.

For the training, we use the Adam optimizer~\citep{kingma2014adam} with a batch size of $16$, a learning rate of $0.002$, and exponential rates of $(\beta_1, \beta_2)=(0.5, 0.999)$. We use a smaller learning rate of $0.0002$ for BERT as suggested in~\citep{devlin2018bert}. The model is trained for 60 epochs.
%
%


\subsection{Notes on Baseline Models}

\textbf{Implementation}: the selected baselines are among the state-of-the-art methods in text-to-image synthesis: DM-GAN\footnote{code available on \url{https://github.com/MinfengZhu/DM-GAN}}~\citep{zhu2019dmgan}, iterative text-to-image synthesis GeNeVA\footnote{code available on \url{https://github.com/Maluuba/GeNeVA}}~\citep{elnouby2019geneva}, attribute-based text-guided image manipulation TA-GAN\footnote{code available on \url{https://github.com/woozzu/tagan}}~\citep{nam2018tagan}. TIRG is a recent cross-modal retrieval\footnote{code available on \url{https://github.com/google/tirg}}~\citep{vo2019tirg} and is adapted for conditional image generation. For these baseline methods, we stick to using their original or adapted official implementation (including their backbone networks and text embeddings) to avoid performance degradation.


\textbf{DM-GAN} is originally used for unconditional text-to-image synthesis and hence has no image input.
To adapt it to our task, we add an image encoder to the model and concatenate the image feature and the text feature as the model input. However, to minimize modification on the architecture, the image feature is squeezed into a vector by using global average pooling. Therefore, significant spatial information of the input image is lost, resulting in low consistency between the generated image and the input image. We add the $\ell_1$ reconstruction loss and find it improves the performance.

\textbf{GeNeVA} is a sequential image synthesis model. We compare it under one-shot generation on the same Abstract scene dataset used in their paper~\citep{elnouby2019geneva}. While GeNeVA is only tested on the ``add'' operation, our method is also verified on other datasets with more diverse and complex text instructions. Applying our method in the sequential generation is non-trivial as it requires the design of extra memory for sequential modeling. Since all baseline methods except GeNeVA do not use memory/state for sequential modeling, we do not evaluate multi-shot generation but leave it as our future work.
\balance
\bibliographystyle{ACM-Reference-Format}
\bibliography{citation}

%% file: math_commands.tex
\usepackage{amsmath,amsfonts,bm}

\def\eqref#1{equation~\ref{#1}}
\def\Eqref#1{Equation~\ref{#1}}

\def\1{\bm{1}}

\DeclareMathAlphabet{\mathsfit}{\encodingdefault}{\sfdefault}{m}{sl}
\SetMathAlphabet{\mathsfit}{bold}{\encodingdefault}{\sfdefault}{bx}{n}



%% file: macros.tex
\def\etal{et~al.}			  
\def\eg{e.g.,~}               
\def\ie{i.e.,~}               
\def\etc{etc}                 
\def\cf{cf.~}                 

\newcommand{\Paragraph}[1]
{\vspace{0.5mm} \noindent \textbf{#1}}
\newlength\paramargin
\newlength\figmargin
\newlength\secmargin
\newlength\figcapmargin

\newcommand{\secref}[1]{Section~\ref{sec:#1}}
\newcommand{\figref}[1]{Figure~\ref{fig:#1}} 
\newcommand{\tabref}[1]{Table~\ref{tab:#1}}

\long\def\ignorethis#1{}

\newenvironment{myitem}%
{\setdefaultleftmargin{2em}{}{}{}{}{} 
 \begin{compactitem}}
{\end{compactitem}}

%% file: 0_abstract.tex
\begin{abstract}
 In recent years, text-guided image manipulation has gained increasing attention in the multimedia and computer vision community. The input to conditional image generation has evolved from image-only to multimodality. In this paper, we study a setting that allows users to edit an image with multiple objects using complex text instructions to add, remove, or change the objects. The inputs of the task are multimodal including (1) a reference image and (2) an instruction in natural language that describes desired modifications to the image. We propose a GAN-based method to tackle this problem.
  The key idea is to treat text as neural operators to locally modify the image feature. We show that the proposed model performs favorably against recent strong baselines on three public datasets. Specifically, it generates images of greater fidelity and semantic relevance, and when used as a image query, leads to better retrieval performance.
\end{abstract}

%% file: 1_introduction.tex
\section{Introduction}

Image synthesis from text has been a highly active research area in the multimedia and computer vision community.
This task is typically set up as a conditional image generation problem where a Generative Adversarial Network (GAN)~\citep{goodfellow2014gan} is learned to generate realistic looking images according to the text description in the format of natural languages~\citep{zhang2018stackgan++,xu2018attngan,zhu2019dmgan,li2019objgan,li2019manigan,nam2018tagan} or scene graphs~\citep{johnson2015sg2im,yikang2019pastegan,li2019controllable,tseng2020retrievegan}, \etc.

In this paper, we study \emph{how to manipulate image content through complex text instruction}.
In this multimodal task, a user is able to apply various changes to a reference image 
by sending text instructions.
For example, \figref{teaser} shows the generated images by the model for three types of instructions: 1) adding a new object at a location, 2) removing an object, and 3) changing the object's attributes (size, shape, color, \etc). This concept was first raised in Schmandt and Hulteen's paper ~\cite{schmandt1982intelligent} and was extended to industrial applications such as PhotoShop through voice commands~\citep{photoshop2017}.

The task studied in this paper is inspired by cross-modal image retrieval -- a cornerstone in many image retrieval tasks such as product search~\citep{kovashka2012whittlesearch,zhao2017memory,guo2018dialog,vo2019tirg,chen2020learning,chen2020image,guo2019fashion}. 
In this retrieval setting~\citep{vo2019tirg,chen2020learning,chen2020image}, users search an image database using a multimodal query that is formed of an image plus some text that describes complex modifications to the input image. 
This retrieval problem is essentially the same as ours except we aim at generating as opposed to retrieving the target image.
Notably, as will be shown in Section~\ref{sec:quant}, the generated image can be used to as a query to retrieve the target images with competitive recall, thereby providing a more explainable search experience that allows users to inspect the search results before the retrieval.

The closest related problem to ours is text-guided image manipulation (\eg~\citep{nam2018tagan,li2019manigan}). However, text instructions in existing works are limited in complexity and diversity as they mainly comprise descriptive attributes, lacking specific actions such as ``add'' or ``remove'' an object. In contrast, the considered text instructions in our paper cover three representative operations ``add'', ``modify'', and ``remove'' and involve adjectives (attributes), verbs (actions) and adverbs (locations) describing the intricate change to one of the objects in the reference image.
Sequential image generation methods~\citep{elnouby2019geneva,cheng2020sequential,chen2019neural}) are also related. For example, GeNeVA~\citep{elnouby2019geneva} generates an image by adding objects to a blank canvas following the step-by-step instructions. Different from ours, these works tackle a different challenge, \ie~temporal modeling of the sequential image generation process.

The main challenge in our problem is how to model the \emph{complex text instructions} for conditional image manipulation. To this end, we propose a simple yet highly effective approach called Text-Instructed Manipulation GAN or \fancynamenospace. The key idea is to treat language as \emph{neural operators} to locally modify the image feature for synthesizing the target image. The text neural operator decomposes the feature modification procedure into two stages: where and how to edit the image feature.
For ``where to edit'', we use attention mechanisms to ground words to a spatial region in the image.
For ``how to edit'', we introduce a text-adaptive network to generate different transformation for varying instructions. 
Since similar instructions perform similar operations, this design allows certain neurons to be shared among similar instructions, while still being able to distinguish among different
operations.

%
%

We conduct extensive experiments on public datasets to demonstrate the three merits of the proposed method. First, it generates high-fidelity images, outperforming recent competitive baselines by a large margin. Second, the user studies confirm that the generated images are more semantically relevant to the target images. Third, the generated image, when used as the query for image-to-image retrieval, leads to not only promising retrieval recalls but also a more explainable search experience that allows users to inspect the results before the search.
In addition, the ablation studies substantiate the performance gain stems from the proposed text operators. Code and models are released at~\url{https://github.com/google/tim-gan}.

%



%% file: 2_related.tex
\vspace{-4mm}
\section{Related Work}
\Paragraph{Conditional generative adversarial networks.}
Generative adversarial networks~GANs~\citep{goodfellow2014gan,mao2017least,arjovsky2017wgan,brock2018biggan,tseng2021regularizing} have made significant progress in recent years.
%
%
Built on the basis of GANs, the \emph{conditional} GAN aims to synthesize the image according to some input context.
The input context can be images~\citep{pix2pix2017,zhu2017cyclegan,lee2020drit++,huang2018munit,mejjati2018attni2i,li2020deepfacepencil}, audio sequences~\citep{lee2019dancing}, human poses~\citep{ma2017pose}, semantic segmentations~\citep{wang2018pix2pixhd,park2019spade,li2019diverse}, \etc.
Among them, text-to-image synthesis~\citep{zhang2018stackgan++,johnson2015sg2im,xu2018attngan,zhu2019dmgan,li2019objgan,yikang2019pastegan,li2019controllablegan,yuan2018text,chen2019neural,li2019storygan} learns a mapping from textual descriptions to images. 
Recently, GeNeVA~\citep{elnouby2019geneva} extended the mapping for iterative image generation in which new objects are added one-by-one to a blank canvas following textual descriptions.
%
%
Different from text-to-image synthesis, the proposed problem takes multimodal inputs, aiming at learning to \emph{manipulate} image content through text instructions.

\Paragraph{Conditional image manipulation.}
The research in this area aims to manipulate image content in a controlled manner.
To enable user-guided manipulation, a variety of frameworks~\citep{zhang2016colorful,zhang2017real,huang2017arbitrary,li2018closed,hung2018learning,portenier2018faceshop,chang2018pairedcyclegan,nam2018tagan,li2019manigan,wu2019gp,zheng2019virtually,zhang2018sparsely,cheng2020sequential,laput2013pixeltone,shinagawa2020interactive,liu2020describe,tseng2020modeling} have been proposed to study different control signals.
For instance, Zhang~\etal~\citep{zhang2017real} and Zou~\etal~\citep{zou2019language} used sparse dots and text respectively to guide the image colorization process.
There are additional works on image manipulation by bounding boxes subsequently refined as semantic masks~\citep{hong2018learning} or code~\citep{mao2019program}.
Numerous image stylization~\citep{huang2017arbitrary,li2018closed} and blending~\citep{hung2018learning,wu2019gp} approaches augment the images by referencing an exemplar image.
Other works include text-guided image inpainting~\citep{zhang2020text,zhang2020text2,lin2020mmfl} which use image caption to inpaint incomplete images.
Closest to ours is the TA-GAN~\citep{nam2018tagan} scheme that takes the image caption as input to describe attributes for conditional image manipulation, followed by~\citep{li2019manigan} and improved by~\cite{liu2020describe}. 
%
In this work, we propose to manipulate the images according to the \emph{complex} text instructions.
Different from the image caption used by the TA-GAN, our instruction is more complex includes 1) three types of operations (``add'', ``remove'', and ``change''); 2) the explicit region information of the modification.

\Paragraph{Multimodal Feature Composition.}
Another related area is multimodal feature composition which has been studied more extensively in other problems such as visual question answering~\citep{kim2016b, noh2016image,chenimage,liang2019focal}, visual reasoning~\citep{johnson2017inferring,santoro2017simple}, image-to-image translation~\citep{zhu2017bicyclegan,lee2020drit++}, \etc. 
Specifically, our method is related to feature-wise modulation, a technique to modulate the features of one source by referencing those from the other. Examples of recent contributions are: text image residual gating~(TIRG)~\citep{vo2019tirg}, feature-wise linear modulation~(FiLM)~\citep{perez2018film}, and feature-wise gating~\citep{ghosh2019interactive}. Among numerous works on multimodal feature composition, this paper compares the closely related methods including a strong feature composition method for image retrieval~\citep{vo2019tirg} and three competitive methods for conditional image generation~\citep{zhu2019dmgan,nam2018tagan,elnouby2019geneva}.

%



%% file: 3_method.tex
\vspace{-3mm}
\section{Methodology}

\begin{figure*}
\vspace{-3mm}
    \centering
    \includegraphics[width=0.98\linewidth]{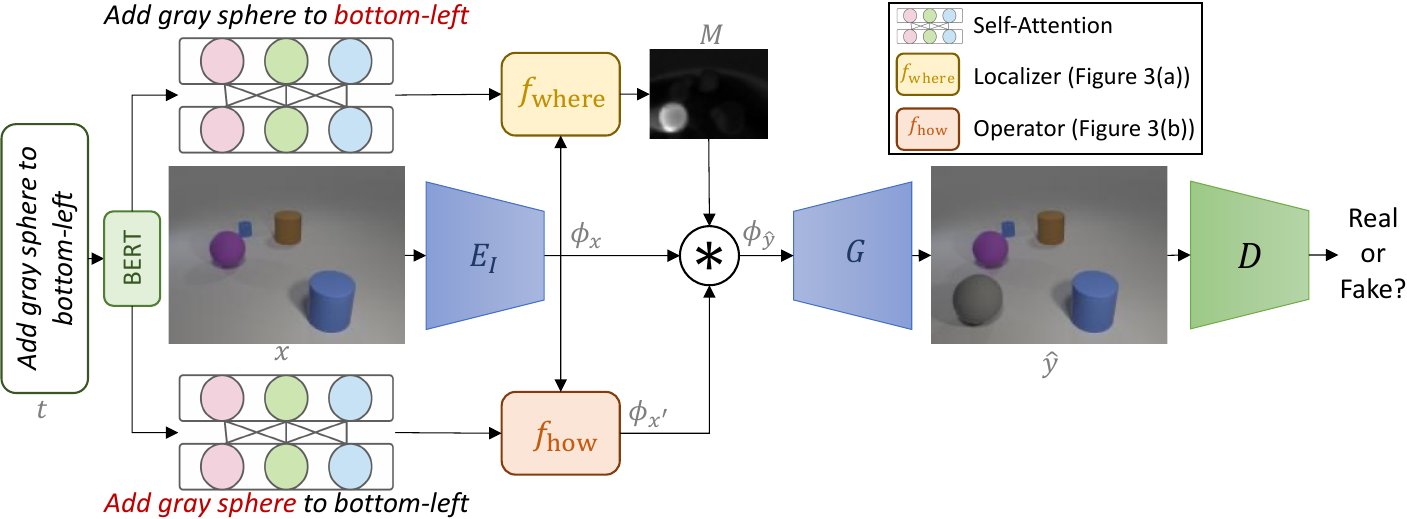}
\vspace{-3mm}
    \caption{\textbf{Method overview.} Given an input image $x$ and a text instruction $t$, the proposed \fancyname first predicts a spatial attention mask $M$ (\emph{where} to edit, \secref{where}) and a text operator $\fhow$ (\emph{how} to edit, \secref{how}).
    The image feature $\phi_{x}$ is then modified by the text operator $\fhow$ on the predicted mask $M$.
    Finally, the edited image $\hat{y}$ is synthesized from the manipulated image feature $\phi_{\hat{y}}$.
    }
    \label{fig:overview}
\vspace{-5mm}
\end{figure*}

Our goal is to manipulate a given reference image according to the modification specified in the input text instruction from one of the three operations: ``add'', ``modify'', and ``remove''.
We approach this problem by modeling instructions as \emph{neural operators} to modify the input image in the feature space. The text operator decomposes this process into two stages: where and how to edit the image feature. Thereafter, the edited feature is used to synthesize the target image by the generator of the GAN model.

An overview of the proposed \fancyname method is illustrated in \figref{overview}.
Given the multimodal input: an image $x$ and a text instruction $t$, our goal is to synthesize an image $\hat{y}$ that is close to the ground-truth target image $y$.
First, we extract the image feature $\phi_x$ and the text features $\phi_t$ where the text encoding comprises two heads producing $\phiwhere$ and $\phihow$ that encodes the \emph{where} and \emph{how} information about the text instruction, respectively.
To indicate the region on the image $x$ to be edited, we predict a spatial attention mask $M$ from $\phiwhere$.
Then, we design a text-adaptive network that embodies a transformation ($\fhow$) for a text embedding ($\phihow$).
Finally, using both the mask $M$ and the embodied function $\fhow$, the input image feature $\phi_x$ is modified into $\phi_{\hat{y}}$, using which the resulting image $\hat{y}$ is generated by the generator $G$.

Formally, the image feature $\phi_x$ is edited by the text operator by:
\begin{align}
\label{eq:overall}
\phiout &= \text{op}_{\text{text}}(\phi_x; t) \\
&=(1-M) \odot \phi_x + M \odot \fop (\phi_x, \phihow; \thetat), \label{eq:overall2}
\end{align}
where $M=\fwhere(\phi_x,\phi_t^\text{where})$ is the learned spatial mask. $\odot$ is element-wise dot product.
The first term is a gated identity establishing the input image feature as a reference to the intended modified feature. 
Although the spatial attention or mask may not be a novel idea in image synthesis~\cite{emami2020spa,lin2021attention,chen2018attention}, we show that disentangling how and where in modification is essential for learning text operators that can be applied at various spatial locations. Our experimental results in Section~\ref{sec:ablations} substantiate this claim.

The second term $\fhow$ embodies the specific computation to obtain the delta modification in the feature space. We introduce a text-adaptive network to execute different transformations for varying text inputs, where each text instruction is identified by a private set of parameters $\thetat$, generated from $\phihow$, and the remaining parameters are shared across all text instructions.

For training, we use the standard conditional GAN objective in the pix2pix \citep{pix2pix2017} model, which consists of an adversarial loss $\mathcal{L}_\mathrm{GAN}$ and an $\ell_1$ reconstruction loss called $\mathcal{L}_{L1}$. The weights to $\mathcal{L}_\mathrm{GAN}$ and $\mathcal{L}_{L1}$ are set to 1 and 10, receptively. In the rest of this section, we will detail the computation of $M$ and $\fhow$.

\begin{figure*}

    \vspace{-4mm}
    \centering
    \begin{subfigure}[t]{\textwidth}
        \centering
        \includegraphics[width=0.9\linewidth]{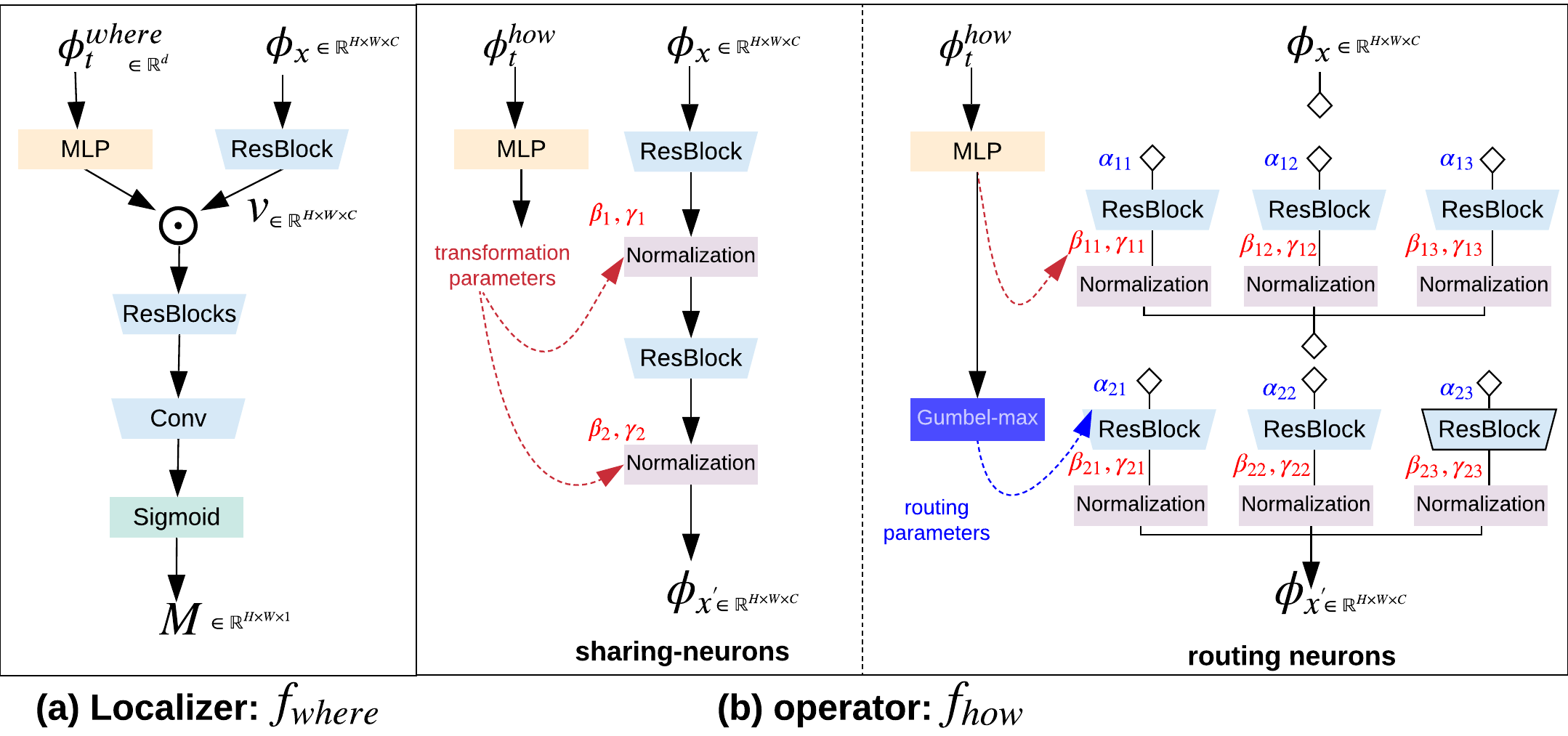}
        \phantomcaption
        \label{fig:other_archa}
    \end{subfigure}
    \begin{subfigure}[t]{0\textwidth} 
        \phantomcaption
    \label{fig:other_archb}   
    \end{subfigure}
    \vspace{-7mm}
    \caption{\label{fig:other_arch}\textbf{\emph{Where} and \emph{how} to edit.} (a) The calculation of spatial mask $M$ from text feature $\phiwhere$ and image feature $\phi_x$.
    (b) The proposed text-adaptive transformation network, parameterized by $(\alpha, \beta, \gamma)$ generated from text feature $\phihow$.
    %
    }
    \vspace{-4mm}
\end{figure*}
\vspace{-2mm}
\subsection{Where to Edit: Spatial Mask}
\label{sec:where}
We use the scaled dot-product self-attention~\citep{vaswani2017attention} to summarize the location-indicative, or locational words, in an instruction. Let $S = [w_1, \cdots, w_l] \in \mathbb{R}^{l \times d_0}$ denote the instruction where $w_i \in \mathbb{R}^{d_0}$ is the 
word embedding~\citep{devlin2018bert} for the $i$-th word. The query, key and value in the attention are computed by:
\vspace{-1mm}
\begin{align}
Q& =  S W_Q, &K& = S W_K, &V& = S W_V
\end{align}
where $W_Q,W_K,W_V\in \mathbb{R}^{d_0\times d}$ are linear weight matrices to learn, and $d$ is the output dimension. After reducing matrix $Q$ to a column vector $\hat{q}$ by average pooling along its first dimension, we obtain the attended text embedding by:
\begin{align}
\phiwhere = V^T\text{softmax}(\frac{K\hat{q}}{\sqrt{d}}),
\end{align}
in which the softmax function encourages higher attention weights over locational words. Likewise, we obtain the text feature $\phihow$ for salient operational words in the instruction (\cf~\figref{editing_b}), computed by a separate self-attention head.

We pass the image feature $\phi_x$ to a convolution block (\ie a ResBlock~\citep{he2016deep}) to get the output $v \in \mathbb{R}^{H \times W \times C}$. The spatial mask is then computed from $\phi_{t}^{\text{where}}$ using image features as the context:
\begin{align}
M &= \fwhere(\phi_x,\phi_t^\text{where}) \nonumber \\
  &= \sigma(W_{m} * (\MLP(\phi_{t}^{\text{where}}) \odot v)) \in [0,1]^{H \times W \times 1}
\end{align}
where $\sigma$ is the sigmoid function, $*$ represents the 2d-convolution product with kernel $W_{m}$ (\cf~\figref{other_archa}). We use two layers of the MLP with the ReLU activation. 

During training, we compute an $\ell_1$ loss to penalize the distance between the predicted mask $M$ and the noisy true mask, and assign it the same weight as the $\mathcal{L}_{L1}$ reconstruction loss.
Note that computing this loss needs no additional supervision as the noisy mask is automatically computed by comparing the difference between the input and ground-truth training images.

\vspace{-2mm}
\subsection{How to Edit: Text-Adaptive Transformation}
\label{sec:how}

Text instructions are not independent. Similar instructions perform similar operations. For instance, ``add a large cylinder'' and ``add a red cylinder'' should perform virtually the same transformation except for the attribute part. Motivated by this idea, we design a text-adaptive network where each text instruction is instantiated by a few private parameters while the rest of the network parameters are being shared across all text instructions. Below, we discuss two types of text-adaptive strategies.

\Paragraph{Sharing-Neurons.} In this strategy, every neuron in the network is shared among all the text instructions. An individual text is identified as a private set of parameters, \ie $\thetat$ in~\Eqref{eq:overall}, calculated from:
\begin{align}
\label{eq:text_op_para}
\thetat =  \MLP(\phiwhere) = \{(\beta_i, \gamma_i) | \gamma_i, \beta_i \in \mathbb{R}^{p}, i \in [1,l]\},
\end{align}
where $l$ is the total number layers of the text-adaptive network illustrated in \figref{other_archb}. Each block consists of a \texttt{conv} layer followed by an instance normalization layer~\citep{ulyanov2017improved}. $p$ is the number of feature channels of each block. From the input feature $\phihow$, an MLP layer is used to generate $\beta$ and $\gamma$ to perform text-specific batch normalization after the ResNet block. Our idea is partially inspired by the style transfer method~\citep{huang2017arbitrary}.

\Paragraph{Routing-Neurons.} We find the above strategy works well in practice but is computationally expensive to scale up. We discuss an alternative strategy to apply text-adaptive transformation inside a routing network~\citep{rosenbaum2017routing} where the text feature is used to dynamically select and execute a sequence of neural blocks (or a path). As a result, we call it routing-neurons strategy.

It is worth noting that our intention is not to compete the routing-neurons strategy with the sharing-neurons strategy because the former often does not lead to further performance gains. Yet, our goal is to show a scalable approach that efficiently increases the learning capacity of text operators, while still allowing certain neurons to be shared among similar instructions.

The text-adaptive network is shown in~\figref{other_archb} which has $l$ layers of $m$ blocks of identical structures. The routing parameter $\alpha_i$ decides to connect or disconnect a block in a layer. A text instruction is hence parameterized by an additional series of $\alpha$:

\vspace{-2mm}
\begin{align}
\label{eq:text_op_para2}
\thetat = \{(\alpha_i, \beta_i, \gamma_i) | \alpha_i \in [0,1]^m, \gamma_i, \beta_i \in \mathbb{R}^{m\times p}, i \in [1,l]\},
\end{align}
\vspace{-2mm}
where $\alpha_i, \beta_i, \gamma_i$ are all generated by the MLPs from $\phihow$.

For efficiency, the path selector $\alpha$ needs to take only discrete values. We employ the Gumbel-Softmax trick \citep{jang2016categorical} to sample a block from a categorical distribution. Let $\pi \in \mathbb{R}_{>0}^{m}$ be the categorical variable with probabilities $P(\alpha=i) \propto \pi_i$, \ie the probability for selecting block $i$. We have:
\begin{align}
\arg\max_i [P(\alpha=i)]= \arg\max_i [g_i + \log \pi_i]=\arg\max_i[\hat{\pi}_i],
\label{eq:gumbel}
\end{align}
where $g_i = -\log(-\log(u_i))$ is a re-parameterization term, and $u_i \sim \text{Uniform}(0,1)$. To make it differentiable, the softmax operation is used to compute $\alpha = \text{softmax}(\hat{\pi}/\tau)$,
where the temperature $\tau$ is set small to encourage $\alpha$ being unimodal. 

\Paragraph{Feature Modification.} Finally. the $\fhow$ function in the text operator (\cf \Eqref{eq:overall2}) is then calculated from:
\begin{align}
\label{eq:text_op_result}
\fhow(\phi_x) &= a^{(l+1)},\\
a^{(i+1)} &= \sum_{j=1}^m \alpha_{ij} (\gamma_{ij} \frac{o_{ij}-\mu(o_{ij})}{\delta(o_{ij})} + \beta_{ij}) & \forall i \in [1,l],\\
a^{(1)} &= \phi_x,
\end{align}
where $o_{ij}$ is the output of the $j$-th \texttt{conv} block in layer $i$. $a^{(i)}$ is the activation of the $i$-th layer. $\delta$ and $\mu$ compute channel-wise mean and variance across spatial dimensions, and are applied at test time unchanged. \Eqref{eq:text_op_result} details the feature modification step for both strategies. In particular, for the sharing-neurons strategy, we fix $j$ to 1 and $\alpha_{i1}=1, \forall i \in [1,l]$ since there is only one path to choose from the network.

%% file: 4_experiments.tex
\section{Experimental Results}

\subsection{Setups}

\Paragraph{Datasets.} \textbf{Clevr:}
CSS dataset~\citep{vo2019tirg} was created for multimodal image retrieval using the Clevr toolkit~\citep{johnson2017clevr}. 
The dataset contains 3-D synthesized images with multiple objects of varying colors, shapes, and sizes. 
Each training sample includes a reference image, a target image and a text specifying the modification from three types ``add'', ``remove'', ``change'' an object. The dataset includes 17K training and 17K test examples.
%
%
%
\textbf{Abstract scene:}
CoDraw~\citep{kim2017codraw} is a dataset built upon Abstract Scene~\citep{zitnick2013bringing} to illustrate a sequence of images of children playing in the park. 
For each sequence, there is a conversation between a Teller and a Drawer. %
The teller gives step-by-step instructions on how to add new content to the current image. Note its text is limited to the ``add'' operation.
To adapt it to our problem, we extract the image and text of a single step.
The dataset consists of 30K training and 8K test examples. 
\textbf{Cityscapes.} 
We create a new dataset of semantic segmentation from the Cityscapes dataset~\cite{Cordts2016Cityscapes}.
There are four types of text modifications: ``add'', ``remove'', ``pull an object closer'', and ``push an object away''. The ground-truth images are manually generated, according to the text instruction, by pasting desired objects onto the image at appropriate positions. 
The dataset consists of 20K training and 3K test examples.


Limited by suitable datasets, related works~\citep{elnouby2019geneva,li2019storygan} were only able to test on synthetic images (\cf more discussions in the supplementary material). In this paper, we extend our method to manipulate semantic segmentation in Cityscapes, and demonstrate the potential of our method for synthesizing RGB images from the modified segmentation mask.


\Paragraph{Baselines.}
We compare with four baseline approaches.
All methods are trained and tested on the same datasets, implemented using their official code or adapted official code. More details about the baseline comparison are discussed in the supplementary material.

\begin{myitem}
\item \textbf{DM-GAN}: The DM-GAN~\citep{zhu2019dmgan} model is a recent text-to-image synthesis framework.
To adapt it to our task, we use our image encoder to extract the image feature and concatenate it with its original text feature as its input signal.

\item \textbf{TIRG-GAN}: TIRG~\citep{vo2019tirg} is a competitive method for the cross-modal image retrieval task. It takes the same input as ours but only produces the image feature for retrieval.
We build a baseline TIRG-GAN based on TIRG by using our image generator $G$ to synthesize the image from the feature produced by the TIRG model.

\item \textbf{TA-GAN}: TA-GAN~\citep{nam2018tagan} learns the mapping between the captions and images.
The image manipulation is conducted by changing the text caption of the image.
Since there is no image caption in our task, we concatenate the pre-trained features of the input image and text instruction as the input caption feature for the TA-GAN model.

\item \textbf{GeNeVA}: GeNeVA~\citep{elnouby2019geneva} learns to generate the image step-by-step according to the text description.
Its main focus is modeling the sequential image generation process. Nevertheless, to adapt it to take the same input as all the other methods, we use it for single-step generation over the real input image.
\end{myitem}
We select the above baseline methods because each of them represents the recent approach for the related problems of \emph{(a)} text-to-image synthesis (DM-GAN), \emph{(b)} multimodal retrieval (TIRG), \emph{(c)} caption-based image manipulation (TA-GAN), and \emph{(d)} sequential image generation (GeNeVA).

\Paragraph{Evaluation Metrics.}
We employ two common metrics: Fr\'echet Inception Distance score (FID)~\cite{fid} and retrieval recall. The former is used to measure the realism of the generated images, and the retrieval recall assesses the semantic relevance between the generated and the true target image. 

To compute the retrieval recall, following~\citep{vo2019tirg,xu2018attngan}, we use the generated image as a query to retrieve the target images in the test set. For simplicity, we compute the cosine similarity between the features of the query and target images where the feature embeddings are obtained by an autoencoder pre-trained on each dataset.

\Paragraph{Implementation Details.}
We implement our model in Pytorch~\citep{paszke2017pytorch}.
For the image encoder $E_I$, we use three down-sampling convolutional layers followed by Instance Normalization with ReLU. 
%
We construct the generator $G$ by using two residual blocks followed by three transposed-convolutional layers with Instance Normalization.
For both $E_I$ and $G$, we use 3x3 kernels and a stride of 2.
%
For the text encoder $E_t$, we use the BERT~\citep{devlin2018bert} model. 
The encoded image has $256$ feature channels and the attended text embedding dimension is $d = 512$. 
By default we use the routing-neurons strategy where the routing network has $l=2$ layers and $m=3$ blocks for each layer.
The parameters in the image encoder $E_i$ and decoder $G$ are initialized by training an image autoencoder for 30 epochs. 
%
Then, we fix $E_i$'s parameters and optimize the other parts of the network in the end-to-end training for 60 epochs.
For training, we use the Adam optimizer~\citep{kingma2014adam} with a batch size of $16$, a learning rate of $0.002$, and exponential rates of $(\beta_1, \beta_2)=(0.5, 0.999)$. 


%

\vspace{-3mm}
\subsection{Main Results}\label{sec:quant}

The main results are shown in~\tabref{quan}, where the Recall@$N$ column indicates the recall ($\times 100$) of the true target image in the top-$N$ retrieved images.
The proposed method performs favorably against all baseline approaches across datasets.
Although DM-GAN appears to generate more realistic images on the Clevr dataset, its retrieval scores are very poor ($<2\%$). This result indicates that it is deceiving to make comparisons only using FID because lower FIDs can be trivially obtained by merely copying the input image without any modifications. 

\begin{figure*}[htbp]
    \centering
    \vspace{-5mm}
    \includegraphics[width=0.9\linewidth]{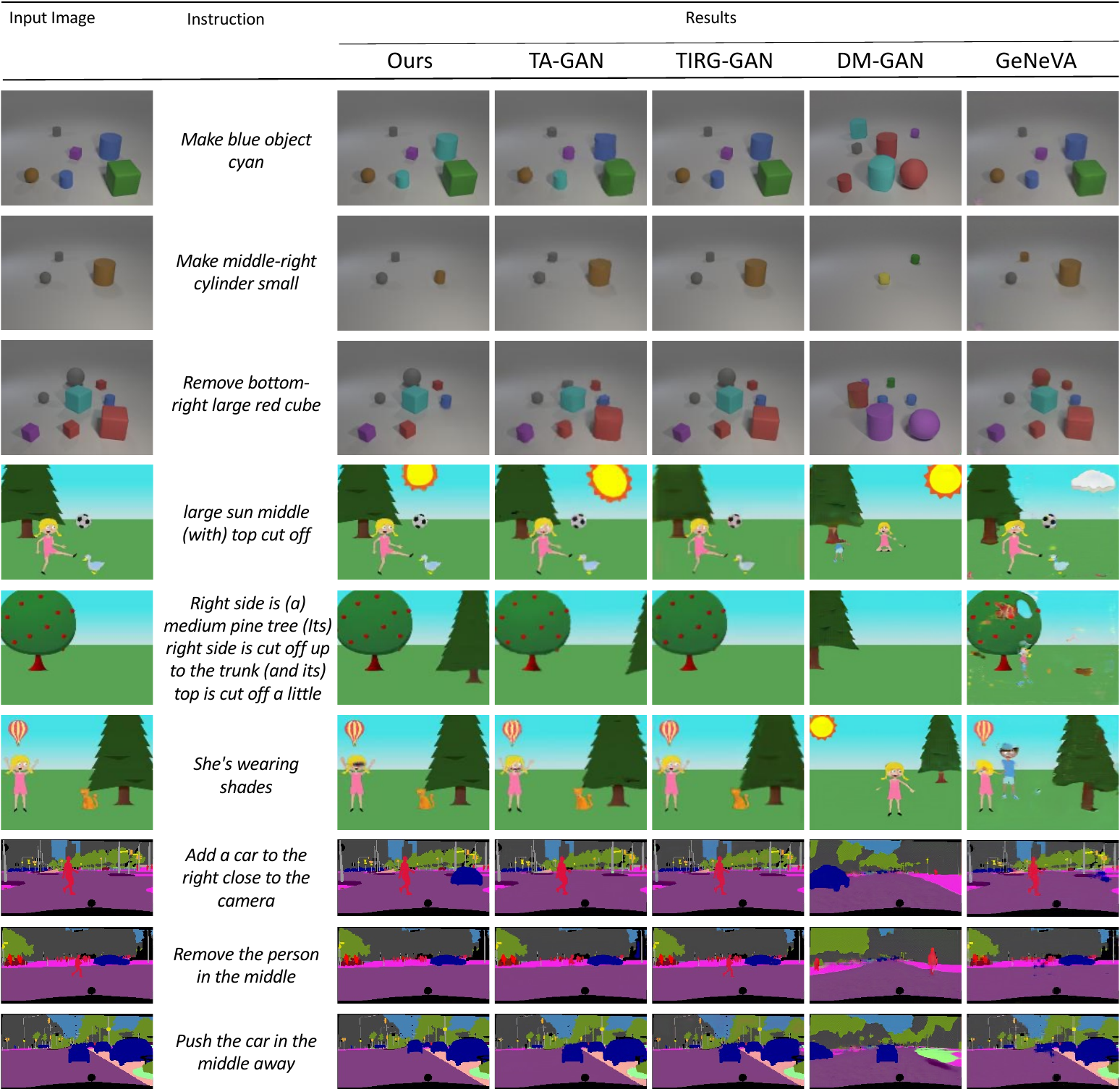}
     \caption{\textbf{Selected generation results.} We show the manipulation results by different approaches on the Clevr (\textit{top}), Abstract scene (\textit{middle}), and Cityscapes (\textit{bottom}) datasets.}
    \label{fig:results}
\end{figure*}

Qualitative results are shown in~\figref{results}. As shown, TA-GAN and TIRG-GAN tend to copy the input images.
DM-GAN often generates random objects following similar input layouts. GeNeVA can make local modifications to images, but often does not follow the text instructions. In contrast, our model generates images guided by the text instructions with greater fidelity and semantic relevance to the true target image.

\begin{figure*}[htbp]
    \vspace{-3mm}
    \centering
    \includegraphics[width=0.9\linewidth]{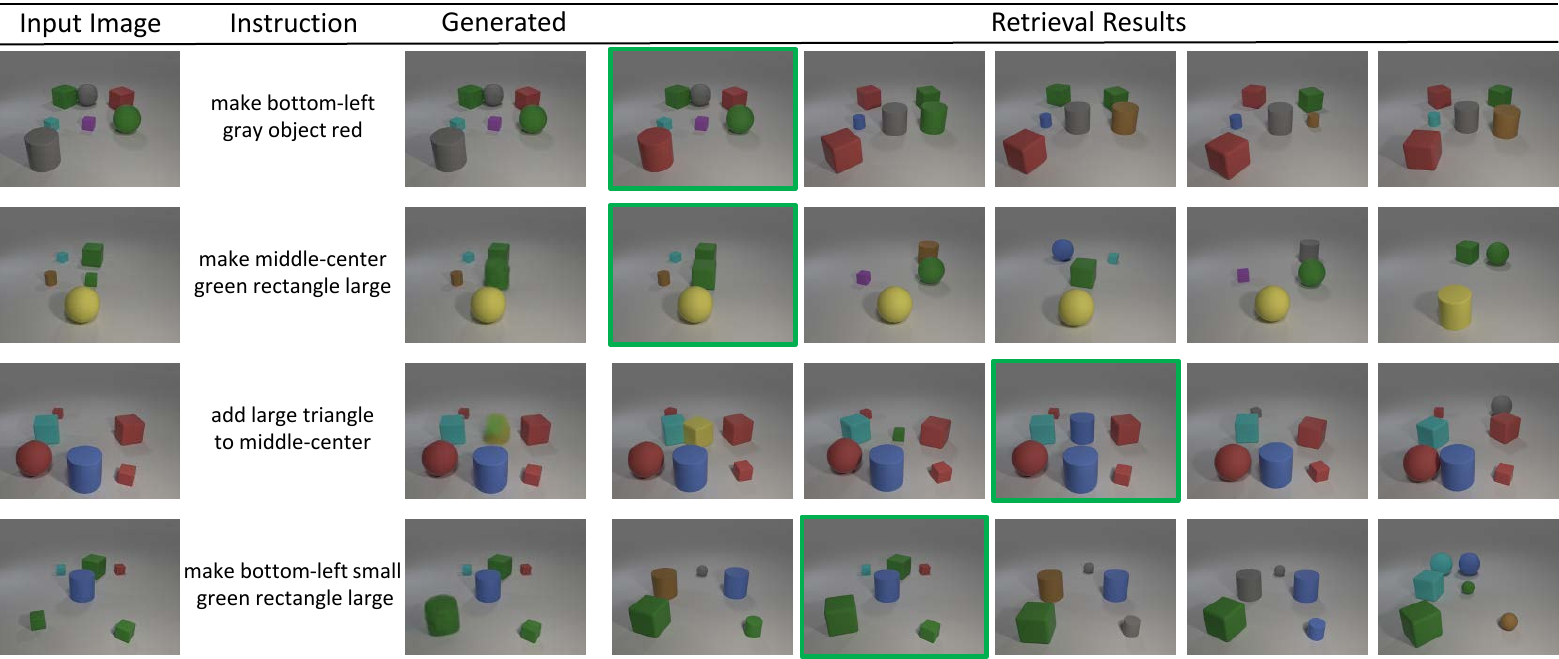}
    \caption{\label{fig:retrieval}Top-5 images retrieved by our generated image (used as the query image). Third column shows the generated (fake) image by our model. Column 4-8 show the top-5 retrieved real images. The true target is highlighted in green.}
\end{figure*}

We use the generated image by our model as a query to retrieve the target image.
\figref{retrieval} shows the top-5 returned images retrieved by our generated image on the Clevr dataset including two successful cases (the first 2 rows in \figref{retrieval}) and two failure cases. There is tangible resemblance between the generated query image and the true target image. This observation is consistent with the quantitative results presented in~\tabref{quan}.

\begin{figure*}[ht]
    \centering
    \vspace{-3mm}
    \begin{subfigure}{0.7\textwidth}
    \includegraphics[width=0.8\linewidth]{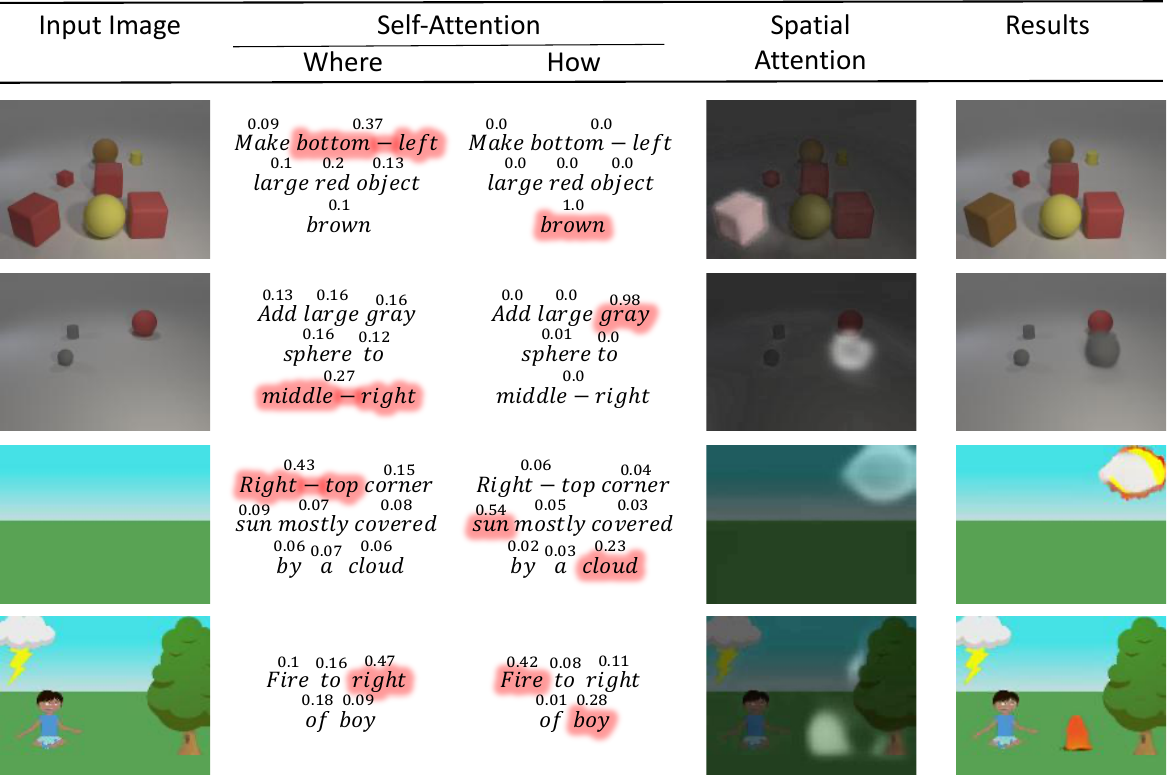}
    \caption{\label{fig:editing_a} Text and spatial attention}
    \end{subfigure}
    \begin{subfigure}{0.28\textwidth}
    \includegraphics[trim=10 0 30 0,width=0.8\linewidth]{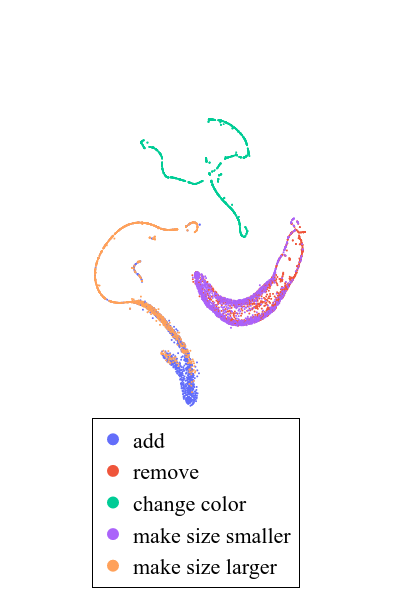}
    \caption{\label{fig:editing_b} Text Operators}
    \end{subfigure}
    \vspace{-3mm}
    \caption{\label{fig:editing}\textbf{Where and how to edit.} (a) Predicted self-attention weights and spatial attention masks.
    The self-attention weights are labeled above each word, and highlighted if the weights are greater than $0.2$. 
    (b) T-SNE visualization of the routing parameters $\alpha$ for various types of text instructions on the Clevr dataset.}
    \vspace{-2mm}
\end{figure*}

\figref{editing} illustrates our intermediate results for where and how to edit, where the learned attention weights for the text and spatial mask are visualized. Generally, the attentions agree with our perception about the task as the self-attentions focus on locational and operational words in the text instruction, respectively, and the spatial attentions capture the intended area for modification.
\input{table/quan.tex}

\begin{figure*}[h]
    \centering
    \includegraphics[width=0.8\linewidth]{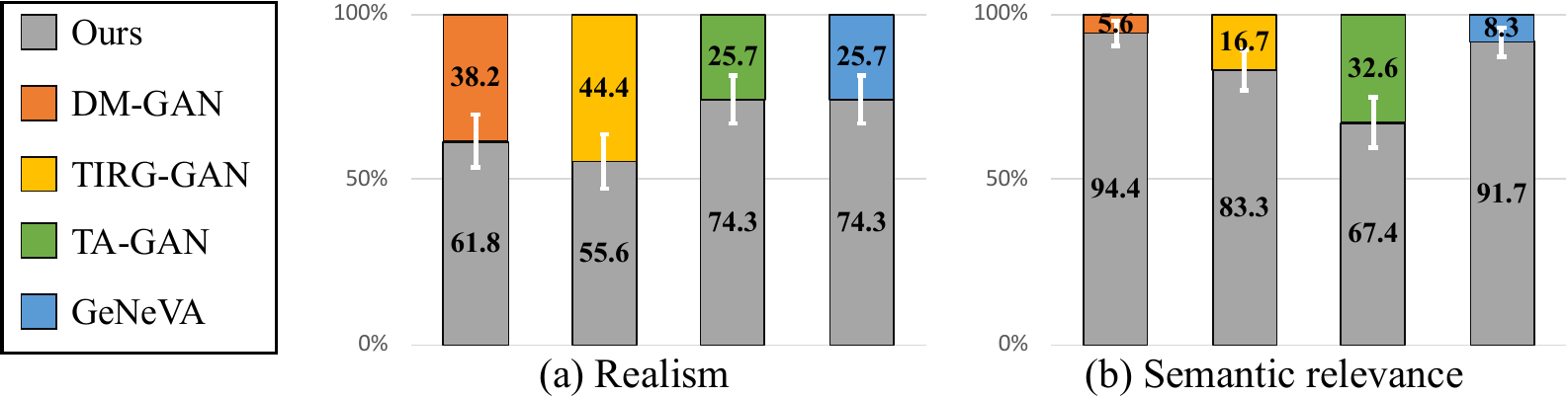}
    \caption{\textbf{User preference studies.} We present manipulated images on the Clevr and abstract scene datasets and ask the users to select the one which (a) is more \emph{realistic} and (b) is more \emph{semantically relevant} to the ground-truth image.}
    \label{fig:userstudy}
\end{figure*}

The above quantitative results, in terms of both FID and retrieval score, substantiate that our method's efficacy in generating high-fidelity and semantically relevant images.


\vspace{-2mm}
\subsection{User preference study}
We conduct two user studies to verify the visual quality and semantic relevance of the generated content. Given a pair of images generated by two compared methods, users are asked to choose \emph{(1)} which one looks more \textit{realistic} while ignoring the input image and text; \emph{(2)} which one is more relevant to the text instruction by comparing the \textit{content} of the generated and the ground-truth image. In total, we collected $960$ answers from $30$ users. 

As shown in \figref{userstudy}, the proposed \fancyname outperforms other methods in both metrics with a statistically significant margin. The above results are consistent with the quantitative results in \tabref{quan}, which validate our method's superior performance in generating not only realistic but also semantically relevant images.

\vspace{-2mm}
\subsection{Ablation studies}\label{sec:ablations}
We test various ablations of our model to validate our design decisions by either leaving the module out from the full model or replacing it with an alternative module.

\Paragraph{Necessity of disentangling how and where to edit.} Our method is built upon a key idea to disentangle how and where to edit. To validate this design, we compare with two entangled text operators in \tabref{ablation1}. The first removes the ``where'' information from the full model by replacing the spatial mask with an identity matrix. The second keeps the spatial mask but discards the ``how'' information by dropping the text-adaptive parameters from the $\fhow$ function. The inferior performance validates the necessity of disentangling how and where to edit. Note that replacing either module leads to worse performance than the baseline methods in \tabref{quan}, which indicates the performance gain is primarily from the proposed text operator as opposed to circumferential factors such as the network backbone or word embedding.
\vspace{-1mm}
\input{table/ablation.tex}

\begin{figure}[h]
    \centering
    \vspace{-4mm}
    \includegraphics[width=.7\linewidth]{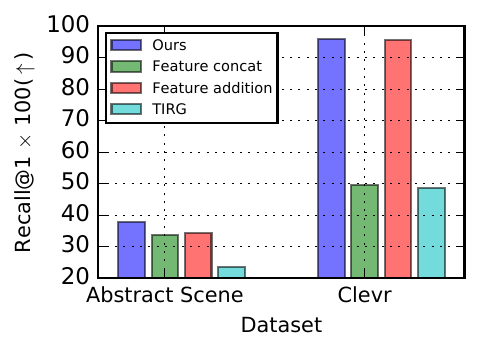}
    \vspace{-4mm}
    \caption{\label{fig:fhow}Comparison between different $\fhow$ functions.}
    \vspace{-4mm}
\end{figure}

\Paragraph{$\fhow$ function.} This experiment compares different designs of the $\fhow$ function. Three alternative models are considered from simple text-and-image feature concatenation, feature addition to the more recent TIRG fusion~\cite{vo2019tirg}. All models use the same \#layers, \#parameters, and spatial mask. \figref{fhow} shows the results. The simple feature addition performs similarly on Clevr but is about 3\% worse than our method on the Abstract Scene dataset.

\Paragraph{With or without routing.} By default, we use the routing-neurons strategy. As discussed in Section~\ref{sec:how}, we do not intent to compete the routing and non-routing strategies because of their similar performances. Nevertheless, we fix the remaining network and study the sharing-neurons (without routing) strategy as well as another softmax routing strategy that computes the continuous routing parameter by the vanilla softmax function. All strategies use two layers and the routing strategies have 3 identical blocks per layer.
\tabref{ablation_flops} lists the results, where the FLOPs and Params column show the number of floating point operations and the number of network parameters of the text operator. We only count FLOPS and Params of the text operator since the remaining parts of the network are shared among strategies.

As discussed in Section~\ref{sec:how}, the routing-neurons strategy leads to no improvements but demonstrates two benefits. First, it allows for scaling up the network parameters at a marginal computational cost. Routing-neurons incurs similar \#FLOPs as Sharing-neurons but inflates \#Params by 3 fold. Second, the routing-neurons strategy enables neural blocks to be shared among similar text operators. \figref{editing_b} shows the t-SNE plot of its routing parameters $\alpha$. It is interesting to find this strategy automatically uncovers the subtle relationship between instructions. For instance, ``add'' and ``make size larger'' operators are closer and share more neural blocks.

\input{table/ablation_flops.tex}

%% file: table/quan.tex
\begin{table*}[htbp]
    \caption{\textbf{Quantitative comparisons.} We use the FID scores to measure the realism of the generated images, and the retrieval score~(RS) to estimate the correspondence to text instructions.}
    \label{tab:quan}
    \setlength{\tabcolsep}{4pt}
    \centering
    \begin{tabular}{@{}l ccc ccc ccc@{}} 
	    \toprule
		\multirow{2}{*}{Method} & \multicolumn{3}{c}{Clevr} & \multicolumn{3}{c}{Abstract scene} & \multicolumn{3}{c}{Cityscape}
		\\  \cmidrule(lr){2-4} \cmidrule(lr){5-7} \cmidrule(lr){8-10} 
		& FID $\downarrow$ & Recall@$1$ $\uparrow$ & Recall@$5$ $\uparrow$ & FID $\downarrow$ & Recall@$1$ $\uparrow$ & Recall@$5$ $\uparrow$ & FID$\downarrow$ & Recall@$1$ $\uparrow$ & Recall@$5$ $\uparrow$ \\
		DM-GAN &$\textbf{27.9}$ &$1.6${\scriptsize$\pm0.1$}&$5.6${\scriptsize$\pm0.1$} &$53.8$ &$2.1${\scriptsize$\pm0.1$} &$6.6${\scriptsize$\pm0.1$}& $18.7$&$4.6${\scriptsize$\pm0.2$} &$15.7${\scriptsize$\pm0.2$} \\
		TIRG-GAN &$34.0$&$\underline{48.5}${\scriptsize$\pm0.2$}&$\underline{68.2}${\scriptsize$\pm0.1$} &$52.7$&$23.5${\scriptsize$\pm0.1$}&$38.8${\scriptsize$\pm0.1$} &$\underline{6.1}$ &$25.0${\scriptsize$\pm0.3$} &$\underline{88.9}${\scriptsize$\pm0.3$}\\
		TA-GAN &$58.8$&$40.8${\scriptsize$\pm0.1$}&$64.1${\scriptsize$\pm0.1$} &$\underline{44.0}$&$\underline{26.9}${\scriptsize$\pm0.2$}&$\underline{46.3}${\scriptsize$\pm0.1$} &$6.7$ &$\underline{36.8}${\scriptsize$\pm0.4$} &$79.8${\scriptsize$\pm0.3$}\\
		GeNeVA &$46.1$&$34.0${\scriptsize$\pm0.1$}&$57.3${\scriptsize$\pm0.1$} &$72.2$&$17.3${\scriptsize$\pm0.2$}&$31.6${\scriptsize$\pm0.2$} & $10.5$&$14.5${\scriptsize$\pm0.4$}&$46.1${\scriptsize$\pm0.3$} \\
		Ours & 
		$\underline{33.0}$ & $\textbf{95.9}${\scriptsize$\pm0.1$} & $\textbf{97.8}${\scriptsize$\pm0.1$} & $\textbf{35.1}$ & $\textbf{35.4}${\scriptsize$\pm0.2$} & $\textbf{58.7}${\scriptsize$\pm0.1$} &$\textbf{5.9}$ &$\textbf{77.2}${\scriptsize$\pm0.4$} &$\textbf{99.9}${\scriptsize$\pm0.1$} \\
		\midrule
		Real images & $17.0$ & $100$ & $100$ & $14.0$ & $100$ & $100$ &$4.4$ & $100$ & $100$ \\
		\bottomrule
    \end{tabular}
    
\end{table*}

%% file: table/ablation.tex

\begin{table}[H]
\vspace{-3mm}
    \caption{Ablation on the disentangled text operator.}
    \vspace{-3mm}
    \label{tab:ablation1}
    \setlength{\tabcolsep}{4pt}
    \centering
    \begin{tabular}{lcccc} 
	    \toprule
		\multicolumn{1}{l}{\multirow{2}{*}{Method}} & \multicolumn{2}{c}{Clevr} & \multicolumn{2}{c}{Abstract scene} \\ 
		& FID $\downarrow$ & \small{Recall@$1$} $\uparrow$ & FID $\downarrow$ & \small{Recall@$1$} $\uparrow$  \\
		\midrule
	    \small{Disentangled text operator} &$\textbf{33.0}$ & $\textbf{95.9}${\scriptsize$\pm0.1$} & $\textbf{35.1}$ & $\textbf{35.4}${\scriptsize$\pm0.2$} \\
        \midrule
		\small{Entangled (no mask)} &$34.8$ & $81.7${\scriptsize$\pm0.1$} & $48.7$ & $28.7${\scriptsize$\pm0.1$} \\
		\small{Entangled (no text-adaptive)} & $45.9$ & $29.9${\scriptsize$\pm0.2$}  & $37.4$ & $33.1${\scriptsize$\pm0.2$} \\
		\bottomrule
    \end{tabular}
\vspace{-5mm}
\end{table}

%% file: table/ablation_flops.tex
\begin{table}[H]
    \caption{Ablation on sharing-neurons and routing-neurons strategies. Only the FLOPS (in Billion) and Params (in Million) of the text operator are compared. }
    \label{tab:ablation_flops}
    \setlength{\tabcolsep}{4pt}
    \centering
    \begin{tabular}{lcccc} 
	    \toprule
		\multicolumn{1}{l}{\multirow{2}{*}{Method}} & \multicolumn{1}{l}{\multirow{2}{*}{FLOPs}} & \multicolumn{1}{l}{\multirow{2}{*}{Params}} & \multicolumn{2}{c}{Clevr} \\ 
		&&& FID $\downarrow$ & \small{Recall@$1$} $\uparrow$  \\
		\midrule
		\small{Sharing-Neurons} &4.08B&4.46M& $33.0$ & $95.8${\scriptsize$\pm0.1$} \\
	    \small{Routing-Neurons} &4.09B&13.91M&$33.0$& $95.9${\scriptsize$\pm0.1$} \\
		\small{Routing (Softmax)} &12.24B&13.91M&$33.0$ & $95.4${\scriptsize$\pm0.1$}  \\
		\bottomrule
    \end{tabular}
\end{table}

%% file: 5_conclusion.tex
\vspace{-4mm}
\section{Conclusion and Future Work}
In this paper, we studied a conditional image generation task that allows users to edit an input image using complex text instructions. We proposed an approach modeling text instructions as neural operators to locally modify the image feature. Our method decomposes ``where'' from ``how'' to apply the modification based on the design of text-adaptive networks. We evaluate our method on one real-world and two synthetic datasets, and obtain promising results with respect to three metrics on image quality, semantic relevance, and retrieval performance. 

In the future, we plan to extend our work on more real-world datasets. Unfortunately, suitable evaluation benchmarks are currently unavailable for real-world RGB images. Therefore, one has to establish an evaluation benchmark of parallel triples of the reference RGB image, target RGB image, and text instruction. Following~\cite{liu2020describe}, we also hope to explore unsupervised training of our model on unpaired text and image data.



